\documentclass{article}

\usepackage[preprint]{neurips_2025}

\usepackage[utf8]{inputenc} 
\usepackage[T1]{fontenc}    
\usepackage{hyperref}       
\usepackage{url}            
\usepackage{booktabs}       
\usepackage{amsfonts}       
\usepackage{nicefrac}       
\usepackage{microtype}      
\usepackage{xcolor}         

\usepackage{colortbl} 
\usepackage{graphicx}
\usepackage{algorithm}
\usepackage{algpseudocode}
\usepackage{amsmath}
\usepackage{amssymb}  
\usepackage{makecell}    
\usepackage{multirow}     
\usepackage{booktabs}    
\usepackage{threeparttable}  
\usepackage{array}      

\usepackage{wrapfig}
\usepackage{multirow}
\usepackage{colortbl}
\definecolor{lightgray}{gray}{0.9}
\definecolor{deepgreen}{RGB}{50,180,50}  
\definecolor{deepred}{RGB}{220,50,50}  
\usepackage{threeparttable}
\usepackage{pifont}   

\usepackage{siunitx}

\usepackage{tikz}
\usepackage[most]{tcolorbox}

\newtcolorbox{examplebox}[1]{
  colback=white,
  colframe=gray!50!black,
  fonttitle=\bfseries,
  title=#1,
  width=\textwidth
}

\newtcolorbox{questionbox}[1]{
  colback=white,
  colframe=gray!50!black,
  fonttitle=\bfseries\color{white},
  colbacktitle=gray!50!black,
  title=#1,
  enhanced,
  frame style={draw=gray!50!black},
  interior style={white},
  drop shadow
}

\newtcolorbox{answerbox}[1]{
  colback=white,
  colframe=gray!50!black,
  fonttitle=\bfseries\color{white},
  colbacktitle=gray!50!black,
  title=#1,
  enhanced,
  frame style={draw=gray!50!black},
  interior style={white},
  drop shadow
}

\definecolor{mypurple}{HTML}{7B3E8B}  
\definecolor{myorange}{HTML}{EE9A49}  
\definecolor{mygreen}{HTML}{2E8B57}   
\definecolor{myred}{HTML}{FF4040}  

\newtcolorbox{correctnessbox}[1]{
  colback=white,
  colframe=mygreen,
  fonttitle=\bfseries\color{white},
  colbacktitle=mygreen,
  title=#1,
  enhanced,
  frame style={draw=mygreen},
  interior style={white},
  drop shadow,
}

\newtcolorbox{incorrectbox}[1]{
  colback=white,
  colframe=myred,
  fonttitle=\bfseries\color{white},
  colbacktitle=myred,
  title=#1,
  enhanced,
  frame style={draw=myred},
  interior style={white},
  drop shadow,
}

\newtcolorbox{tokenbox}[1]{
  colback=white,
  colframe=myorange,
  fonttitle=\bfseries\color{white},
  colbacktitle=myorange,
  title=#1,
  enhanced,
  frame style={draw=myorange},
  interior style={white},
  drop shadow,
}

\newtcolorbox{reflectionbox}[1]{
  colback=white,
  colframe=mypurple,
  fonttitle=\bfseries\color{white},
  colbacktitle=mypurple,
  title=#1,
  enhanced,
  frame style={draw=mypurple},
  interior style={white},
  drop shadow,
}

\newtcolorbox{responsebox}[1]{
  colback=white,
  colframe=gray!50!black,
  fonttitle=\bfseries\color{white},
  colbacktitle=gray!50!black,
  title=#1,
  enhanced,
  frame style={draw=gray!50!black},
  interior style={white},
  drop shadow
}

\newtcolorbox{thinkingbox}[1]{
  colback=gray!20,
  colframe=gray!30,
  fonttitle=\bfseries,
  title=#1
}

\newtcolorbox{takeawaybox}[1]{
  enhanced,
  colback=white,
  colframe=black,
  arc=10pt,
  title={#1},
  fonttitle=\large\bfseries,
  colbacktitle=black,
  coltitle=white,
  attach boxed title to top center={yshift=-3mm},
  boxed title style={
    size=small,
    arc=5pt,
    rounded corners
  }
}

\usepackage{amssymb}
\usepackage{amsmath}
\usepackage{multirow}
\usepackage{amsthm}  

\definecolor{lightgray}{gray}{0.9} 

\DeclareMathOperator*{\argmax}{arg\,max}  

\title{AdapThink: Adaptive Thinking Preferences for Reasoning Language Model}

\author{%
  Xu Wan\\
  Zhejiang University\\
  \texttt{wanxu@zju.edu.cn} \\
  \And
  Wei Wang\\
  Alibaba DAMO Academy\\
  \texttt{richard.wang1@unimelb.edu.au} \\
  \And
  Wenyue Xu\\
  Tongji University\\
  \texttt{2111141@tongji.edu.cn} \\
  \And
  Wotao Yin\\
  Alibaba DAMO Academy\\
  \texttt{wotao.yin@alibaba-inc.com} \\
  \And
  Jie Song\\
  Peking University\\
  \texttt{songjie@coe.pku.edu.cn} \\
  \And
  Mingyang Sun\thanks{Coressponding author.}\\
  Peking University\\
  \texttt{smy@pku.edu.cn} \\
}

\begin{document}

\maketitle

\begin{abstract}
  Reinforcement Learning (RL)-based post-training has significantly advanced the complex reasoning capabilities of language models, fostering sophisticated self-reflection processes. However, this ``slow thinking'' paradigm presents a critical challenge to reasoning efficiency: models may expend excessive computation on simple questions and shift reasoning prematurely for complex ones. Previous mechanisms typically rely on static length budgets or predefined rules, lacking the adaptability for varying question complexities and models' evolving capabilities. To this end, we propose \textbf{\textit{AdapThink}}, an \textbf{\textit{adap}}tive post-training framework designed to induce more efficient \textbf{\textit{think}}ing while maintaining the performance of reasoning language models. Specifically, AdapThink incorporates two key mechanisms: 1) A group-relative reward function that leverages model confidence and response's characteristic to dynamically adjust the preference of reflection-related transition words without resorting to a fixed length preference. 2) A diversity-aware sampling mechanism that balances the training group's solution accuracy with reasoning diversity via an entropy-guided score. Experiments on several mathematical reasoning datasets with DeepSeek-distilled models demonstrate AdapThink's advantages in enabling adaptive reasoning patterns and mitigating the inefficiencies.
  \end{abstract}

\section{Introduction}

Recent breakthroughs in large language models such as OpenAI's o1 \cite{openai2024learning} and DeepSeek R1 \cite{guo2025deepseek} have demonstrated that reinforcement learning (RL)-based post-training methods can substantially improve their reasoning capabilities. This enhancement is primarily attributed to the emergence of models' sophisticated self-reflection behaviors~\cite{kumar2025llm, kazemnejad2024vineppo}. However, recent research has highlighted a significant inefficiency associated with this ``\textit{slow thinking}'' pattern \cite{muennighoff2025s1simpletesttimescaling, han2025tokenbudgetawarellmreasoning, chen2025think23overthinkingo1like, wang2025thoughtsplaceunderthinkingo1like, aggarwal2025l1controllinglongreasoning, shen2025dastdifficultyadaptiveslowthinkinglarge}. Reasoning models frequently \textit{overthink} simple problems, thereby spending unnecessary computational resources, while conversely \textit{underthinking} complex challenges, leading to incomplete reasoning and incorrect answers.
When presented with a simple problem (see Appendix \ref{app::example}), a model can reach a correct answer with merely 479 tokens. Yet, its self-validation mechanisms---marked by phrases such as ``Verify'' and ``Wait''---triggered unnecessary reflections, resulting in more than a quadrupled token consumption. In contrast, when attempting a complex problem, the same model exhibited frequent and unproductive shifts, marked by phrases ``Alternatively'' and ``Another.'' Upon reaching its token limit, the model stopped and arrived at an incorrect answer.

Therefore, an ideal chain of thought (CoT) would be capable of adjusting its self-reflection frequency and depth, adaptive to the problem difficulty and its level of confidence. Empirically, \cite{ma2025rethinking, xie2025logic} have demonstrated that employing reflection vocabularies does \emph{not always} guarantee mathematical reasoning improvement. Drawing upon this insight, recent works explored direct control via modifying the input prompt \cite{jin2024impactreasoningsteplength, muennighoff2025s1simpletesttimescaling, liu2025adaptivestepautomaticallydividingreasoning, han2025tokenbudgetawarellmreasoning, chen2025think23overthinkingo1like} or providing indirect reasoning-length rewards \cite{wang2025thoughtsplaceunderthinkingo1like, aggarwal2025l1controllinglongreasoning, shen2025dastdifficultyadaptiveslowthinkinglarge}. 
A common limitation of these approaches is using token budgets governed by rules or offering rewards for adhering to a special length budget, overlooking the critical impact of the changes in models' capabilities and length preferences. Intuitively, models with limited reasoning abilities benefit from more extended CoT patterns, as redundant self-reflection could serendipitously contribute to reaching correct solutions. In contrast, high-performing models should aim to minimize token consumption to prevent overthinking and to maintain efficient CoT reasoning.

Motivated by these insights and our observations in Section \ref{sec::observation}, we propose \textbf{AdapThink}, an \emph{RL post-training framework} for reasoning models. Our post-training empowers models to adjust their preference for reasoning depth, which is adaptive to their current operating capabilities. Instead of directly limiting the budget of reasoning length, our work adjusts the length preference through analyzing the distribution of diverse reasoning patterns observed in \emph{groups of generated samples}. We find that {reasoning length alone does not directly determine model reasoning performance. Instead, strategically regulating reasoning depth is a more effective way to mitigate overthinking and underthinking.} Specifically, our main contributions are:

\begin{enumerate}
\item  First, we introduce \emph{a group-relative reward function}, a novel mechanism designed to adjust the model's current reasoning preferences. With it, the model is trained to determine the appropriate preferences for reflection based on intra-group accuracy of generated responses, while measuring reasoning efficiency \textit{quantitatively} through counting key transition words that appear in these groups of training samples.

\item Furthermore, we propose a \emph{diversity-aware sampling mechanism} to balance between achieving high reasoning accuracy and fostering rich diversity within the groups of training samples. To achieve this, we begin with an oversampling of reasoning instances. We then employ carefully defined \textit{diversity metrics} to evaluate both the final answers and the intermediate steps of these instances, before applying diversity-aware downsampling to curate and enhance the overall quality of the instances used for RL post-training.

\item By post-training the DeepSeek-distilled Qwen model subject to a context length limit of only \emph{2K tokens}, our method achieves superior performance compared to several length-control baselines when tested within an 8K-token limit, demonstrating AdapThink's effectiveness in developing powerful and efficient CoT models across multiple reasoning benchmarks.
\end{enumerate}

\section{Related Work}

\paragraph{Induce Longer CoT} For inducing longer reasoning length, several works \cite{weng2023large,miao2023selfcheck,saunders2022self,renze2024self,jin2024impactreasoningsteplength} have encouraged models to engage in deeper thinking through natural language feedback. For zero-shot CoT, \cite{weng2023large, miao2023selfcheck} stimulate model self-reflection by performing backward verification or multiple response voting. For few-shot CoT with demonstrations, \cite{jin2024impactreasoningsteplength} introduced five general standardized patterns to induce models to simulate human thinking and reshape the CoT.
However, these one-size-fits-all approaches ignore the diversity of problem-solving paths, only have a monotonous processing mode. Similarly, \cite{muennighoff2025s1simpletesttimescaling} designed several budget forcing mechanisms to increase guidance in CoT, such as appending ``Wait'' multiple times when the model tries to end, forcing it to double-check. Likewise, \cite{shen2025satorireinforcementlearningchainofactionthought} introduced special meta-action markers like \texttt{<|continue|>}, \texttt{<|reflect|>}, and \texttt{<|explore|>}, enabling the model to restart from intermediate steps, lengthen responses, and correct errors.

\paragraph{Induce Shorter CoT} Simple prompt methods are also effective for encouraging a shorter CoT \cite{nayab2024concise,xu2025chain,renze2024benefits}. For example, \cite{muennighoff2025s1simpletesttimescaling} added “Final answer” to terminate the model’s thinking process. Besides, \cite{jin2024impactreasoningsteplength, kang2025c3ot} used stronger models to compress long CoTs semantically into shorter ones. 
\cite{yeo2025demystifyinglongchainofthoughtreasoning, shen2025dastdifficultyadaptiveslowthinkinglarge, aggarwal2025l1controllinglongreasoning, yu2025dapoopensourcellmreinforcement} design different length-budget signals from the perspective of rule-based reward design, encouraging models to balance accuracy and token efficiency during the thinking process. Although \cite{shen2025dastdifficultyadaptiveslowthinkinglarge} already takes problem complexity and model confidence into account for the budget, its adoption of a uniformly shorter response preference is less suitable for models with weaker reasoning abilities. Moreover, while \cite{yu2025dapoopensourcellmreinforcement,zhang2025srpo} uses dynamic sampling to avoid the zero-advantage issue in group-relative advantage calculation, its sampling strategy focuses solely on outcome diversity metrics and neglects intra-group reasoning process diversity.

Moreover, some studies have further focused on addressing overthinking and underthinking in long CoTs. \cite{wang2025thoughtsplaceunderthinkingo1like,sui2025stop} introduced thought switching penalties to influence the token decoding probability distribution early in CoT generation, reducing initial thought-switching. Similarly, \cite{chen2025think23overthinkingo1like} presented an efficiency metric to evaluate each token’s contribution to accuracy and used length preference optimization to achieve more efficient CoT patterns. However, both approaches depend on auxiliary judgments from more powerful reasoning models, making the performance improvements inherently constrained by the capabilities of the reference models.

\section{Observations}
\label{sec::observation}
To investigate potential overthinking and underthinking issues in current language models, we first conducted a comprehensive analysis of the generation patterns from DeepSeek-R1-Distill-Qwen-1.5B and DeepSeek-R1-Distill-Qwen-7B models \cite{guo2025deepseek}. We examined the CoT generation efficiency from three perspectives:

\begin{itemize}
    \item \textbf{Token Length Distribution}: Following the inference settings in \cite{aggarwal2025l1controllinglongreasoning}, we set the maximum token limit to 8,192 during inference and segmented the model's response range into four equal intervals to analyze the distribution patterns.
    \item \textbf{``Pause-Validation'' Words Distribution}: We analyzed the frequency of transition words in the responses, focusing on four key indicators of ``Pause-Validation'': ``\textit{wait}'', ``\textit{hold on}'', ``\textit{check}'', and ``\textit{verify}''. The distribution was categorized into four bins with an interval of 3.
    \item \textbf{``Branch-Extension'' Words Distribution}: Similarly, we identified four representative markers of ``Branch-Extension'': ``\textit{alternatively}'', ``\textit{however}'', ``\textit{another}'', and ``\textit{instead}''. Their occurrences were also categorized into four bins with an interval of 3.
\end{itemize}

\begin{figure}[t]
\centering
\includegraphics[width=\columnwidth]{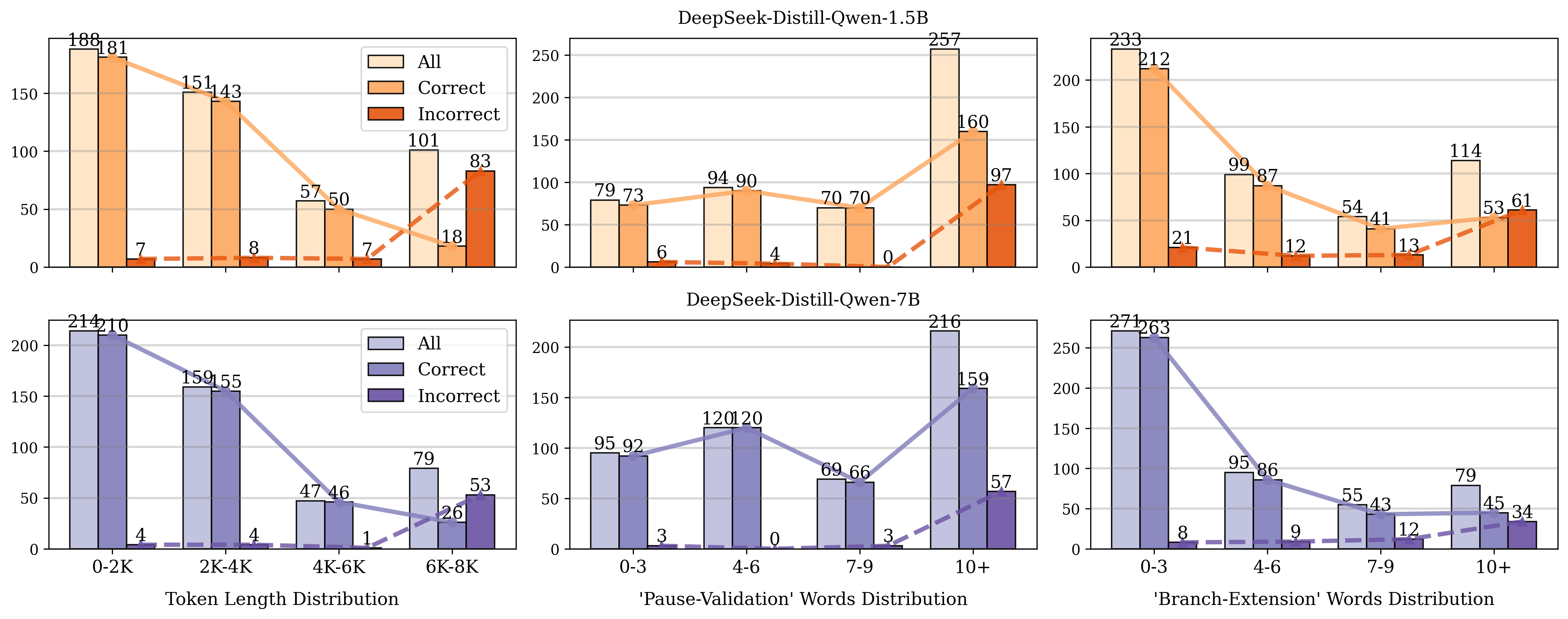}
\caption{Distribution analysis for DeepSeek-Distill-Qwen 1.5B (top), 7B (bottom) on MATH-500. X-axes: token ranges (left), word frequency intervals (middle/right); Y-axes: sample counts.}
\label{fig::observation}
\end{figure}

As shown in Figure \ref{fig::observation}, incorrect answers typically correlate with higher token consumption and increased usage of both ``Pause-Validation'' and ``Branch-Extension'' words. This suggests that when confronting challenging problems, models tend to exhibit uncertain behavior through excessive alternative explorations and repetitive self-verifications. Intriguingly, for correct answers, we observed that the responses predominantly fall within lower token ranges and show lower frequencies of ``Branch-Extension'' word occurrences. Notably, while correct answers demonstrate reduced exploration of alternative paths, they maintain consistent levels of ``Pause-Validation'' expressions, suggesting a potential distinction between the roles of different transition words in the reasoning process. These observations motivate us to explore effective ways to achieve more efficient reasoning while maintaining performance, with particular attention to the distinct roles of different transition words in the reasoning process.

\section{Methdology}
\label{sec::methodology}
In this section, we introduce the proposed post-training framework AdapThink in detail. Unlike previous approaches that \emph{directly impose length restrictions} or set \textit{token budgets} \cite{shen2025dastdifficultyadaptiveslowthinkinglarge, yeo2025demystifyinglongchainofthoughtreasoning, aggarwal2025l1controllinglongreasoning, muennighoff2025s1simpletesttimescaling}, we propose to modulate the reasoning behavior by regulating the preference of reflection words based on group-level response characteristics. To this end, AdapThink introduces two key innovations: (1) a group-relative reward function for controlling reasoning preferences, and (2) a diversity-aware sampling mechanism for ensuring intra-group diversity during training. The pseudo-code of AdapThink can be found in Appendix \ref{app::pseudo-code}.


\subsection{Group-relative Reasoning Preference Reward}

We begin with a pre-trained reasoning language model $\pi_{\theta}$ and a dataset $\mathcal{D} = \{(x^k, y^k)\}_{k=1}^{N}$, where each instance contains the input prompt $x$ and the final answer $y$. For each $x \in \mathcal{D}$, $\pi_{\theta}$ performs reasoning to generate $|\mathcal{G}|$ samples $\mathcal{G} := \{y_i\}_{i=1}^{|\mathcal{G}|}$. Next, we analyze these samples from two perspectives:
    
\begin{itemize}
    \item \textbf{Model Confidence.} We represent the average correctness of the model's $|\mathcal{G}|$ samples within a group as the model confidence, denoted as $\varphi$. 
    \begin{equation}
        \varphi = \frac{1}{|\mathcal{G}|}\sum_{i = 1}^{|\mathcal{G}|} \mathbb{I}(y_i = y),
        \label{eq::confidence}
    \end{equation}
    where $\mathbb{I}(\cdot)$ is the indicator function . A larger $\varphi$ reflects more reliable reasoning patterns for the current question $x$. We set $\varphi_{\text{low}}$ and $\varphi_{\text{high}}$ as the boundaries for low and high confidence levels. $\varphi$ directly influences the p\textit{reference direction} of the AdapThink's reasoning preference reward.
    \item \textbf{Response Characteristics.} Given the distinct reasoning patterns observed between correct and incorrect answers, we partition the model's responses into two groups based on their correctness: $\mathcal{G_T}$ for correct answers and $\mathcal{G_F}$ for incorrect ones. The distribution of transition words within these groups determines the \textit{specific values} of the reasoning preference reward.
\end{itemize}

Then, based on the reflection words analysis in Section \ref{sec::observation}, we aim to control the usage of ``Branch-Extension'' transition words while promoting efficient output generation. Specifically, we introduce three reward components to guide the reasoning process: token consumption reward $\lambda_l$ that quantifies the response length, output completion reward $\lambda_o$ that tracks the presence of answer completion marker in the \texttt{</think>} tag, and ``Branch-Extension'' transition words reward $\lambda_b$ that measures the numbers of four switch words. Each component reward $\lambda_*$ is calculated as the normalized deviation from its corresponding group mean:
\begin{equation}
\lambda_*(y_j) = \frac{r_*(y_j)}{\mu_*(\mathcal{G}(y_j))} - 1, \quad * \in \{l, o, b\}
\label{eq::component_reward}
\end{equation}
where $\mathcal{G}(y_j)$ maps response $y_j$ to either $\mathcal{G_T}$ or $\mathcal{G_F}$ depending on whether $y_i$ is correct or not, and $\mu_*(\mathcal{G})$ represents the mean measurement within group $\mathcal{G}$.

Then, we introduce the group-relative reasoning preference reward (GRPR) to adaptively enable different reward components based on the model's confidence $\varphi$ for the current question:
\begin{equation}
r(x, \mathcal{G}, \theta) = \text{clip}\Big(|\omega(\varphi)|(\lambda_o - \lambda_l) + \mathbb{I}(\omega(\varphi) < 0)\omega(\varphi)\lambda_b, r_{\text{min}}, r_{\text{max}}\Big)
\label{eq::reason_reward}
\end{equation}
where the clip function bounds the reward within $[r_{\text{min}}, r_\text{max}]$ to ensure stable training signals. Here, $\omega(\varphi)$ is a cosine interpolation function that smoothly transitions between reward preferences at confidence boundaries:
\begin{equation}
\omega(\varphi) = \begin{cases}
+1 & \text{if } \varphi \leq \varphi_{\text{low}} \\
\cos(\frac{\varphi - \varphi_{\text{low}}}{\varphi_{\text{high}} - \varphi_{\text{low}}} \pi) & \text{if } \varphi_{\text{low}} < \varphi < \varphi_{\text{high}} \\
-1 & \text{if } \varphi \geq \varphi_{\text{high}}
\end{cases}
\label{eq::weight}
\end{equation}

For low-confidence cases ($\omega(\varphi) > 0$), the GRPR reward emphasizes output completion while controlling length; for high-confidence cases ($\omega(\varphi) < 0$), it additionally controls the branch-extension exploration.

\begin{figure}[t]
\centering
\includegraphics[width=\columnwidth]{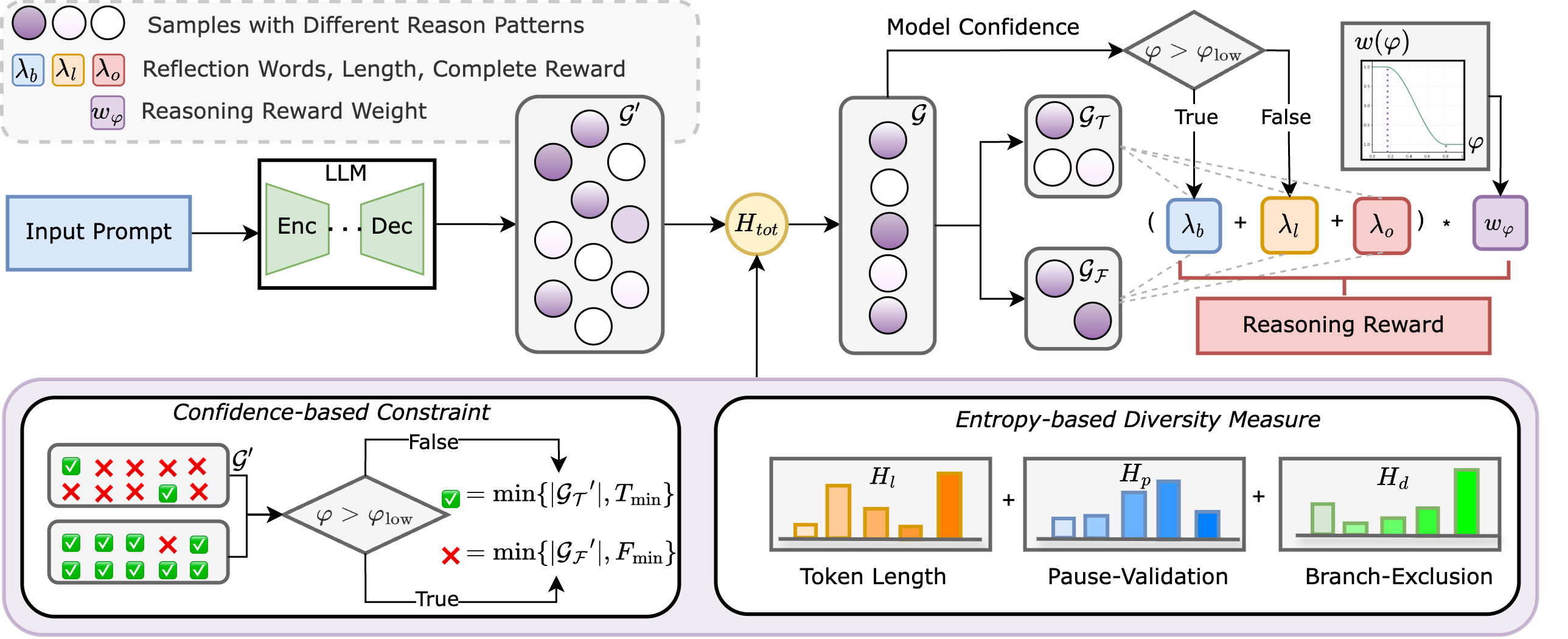}
\caption{The AdapThink framework architecture, featuring confidence-guided sample selection and entropy-based diversity measurement for adaptive reasoning reward computation.}
\label{fig::framework}
\end{figure}

\subsection{Diversity-aware Sampling}
As shown in Equation \ref{eq::reason_reward}, since GRPR incorporates group-specific distribution patterns from $\mathcal{G_T}$ and $\mathcal{G_F}$ into its reward computation, the diversity of reason patterns within each group significantly influences the effectiveness of GRPR-based post-training. While previous studies have addressed the zero-advantage problem through dynamic upsampling strategies \cite{yu2025dapoopensourcellmreinforcement} to avoid homogeneous output labels within groups, they haven't considered the diversity of reasoning processes.

To address this limitation, we propose a two-stage sampling strategy that enhances reasoning diversity while maintaining group balance:

\textbf{Stage 1: Oversampling and Diversity Measurement}
First, we oversample by a factor of $K$ to expand the sample pool to $\mathcal{G}'$ candidates, allowing exploration of diverse reasoning styles within each group. To evaluate sample diversity, we introduce entropy-based measures $H_l$, $H_p$, and $H_b$ for token length,``Pause-Validation'' words and ``Branch-Extension'' words, respectively:
\begin{equation}
    H_* (\mathcal{G}') = -\frac{1}{\log |\mathcal{G}'|}\sum_{s \in \mathcal{S}} \frac{|s|}{|\mathcal{G}'|} \log \frac{|s|}{|\mathcal{G}'|}, \quad * \in \{l, b, p\}
\end{equation}

where we partition each component's measurement range into four equal intervals, forming bins $\mathcal{S}$. The normalized entropy ensures $H_* \in [0,1]$, with higher values indicating a more uniform distribution for reflection words count. The overall diversity score combines these dimensions:

\begin{equation}
    H_{\text{tot}} = \alpha_l H_l + \alpha_p H_p + \alpha_b H_b
\end{equation}
where $\alpha_l$, $\alpha_p$ and $\alpha_b$ denote the weights for length, ``Pause-Validation'' words and ``Branch-Extension'' words diversity score, respectively.

\textbf{Stage 2: Confidence-constrained Downsampling}
To enhance RL training efficiency while maintaining sample diversity, we select $|\mathcal{G}|$ samples from the oversampled pool $\mathcal{G}'$, which consists of correct answers $\mathcal{G_T}'$ and incorrect answers $\mathcal{G_F}'$. To prevent model bias towards a single output label, we calculate the minimum number of samples required from each category as shown in Equation \ref{eq::conditional_greedy}:

\begin{equation}
    T = \min\{|\mathcal{G_T}'|, T_{\min}\}, \quad F = \min\{|\mathcal{G_F}'|, F_{\min}\}
\end{equation}

where $T_{\min}$ and $F_{\min}$ represent the minimum requirements for correct and incorrect samples, respectively. Based on the model's confidence level $\varphi$, we adapt our selection strategy to prioritize either $T$ correct or $F$ incorrect samples while maximizing reasoning diversity. 
The final sample selection process adapts to the model's confidence level $\varphi$, as formulated in Equation \ref{eq::conditional_greedy}:

\begin{equation}
\mathcal{G} = \begin{cases}
\argmax\limits_{\mathcal{G}_c \subseteq \mathcal{G_T}', |\mathcal{G}_c|=T} H_{\text{tot}}(\mathcal{G}_c) \cup \argmax\limits_{\mathcal{G}_s \subseteq \mathcal{G}'_{\text{remain}}, |\mathcal{G}_s|=|\mathcal{G}|-T} H_{\text{tot}}(\mathcal{G}_s) & \text{if } \varphi \leq \varphi_{\text{low}} \\
\argmax\limits_{\mathcal{G}_c \subseteq \mathcal{G_F}', |\mathcal{G}_c|=F} H_{\text{tot}}(\mathcal{G}_c) \cup \argmax\limits_{\mathcal{G}_s \subseteq \mathcal{G}'_{\text{remain}}, |\mathcal{G}_s|=|\mathcal{G}|-F} H_{\text{tot}}(\mathcal{G}_s) & \text{otherwise}
\end{cases}
\label{eq::conditional_greedy}
\end{equation}

where $\mathcal{G_T}'$ and $\mathcal{G_F}'$ denote the pools of correct and incorrect answers from oversampling, respectively. When model confidence is low ($\varphi \leq \varphi_{\text{low}}$), the algorithm first selects $T$ correct samples $\mathcal{G}_c$ from $\mathcal{G_T}'$ that maximize diversity; otherwise, it prioritizes selecting $F$ incorrect samples from $\mathcal{G_F}'$. The remaining $\mathcal{G}_s$ training samples are then selected from the candidate pool $\mathcal{G}'_{\text{remain}}$  to reach the target size $|\mathcal{G}|$, while maintaining maximum diversity measured by $H_{\text{tot}}$.

\section{Experimental Setup}
\label{sec::exp_set}
\paragraph{Datasets and Pretrained LLMs} We conduct experiments on a curated lightweight mathematics dataset that spans various difficulty levels. This dataset combines queations from \textit{DeepScaleR-Preview-Dataset} \cite{deepscaler2025}, including about 5K question-answer pairs sampled from AIME (1984-2023), AMC (prior to 2023), and MATH training sets. For baseline models, we employ \textit{DeepSeek-R1-Distill-Qwen-1.5B}, which has demonstrated competitive performance against GPT-4-0513 and QwQ-32B-Preview on various mathematical reasoning benchmarks \cite{guo2025deepseek}.

\paragraph{Baselines} We conduct post-training on DeepSeek-R1-Distill-Qwen-1.5B using Group Relative Policy Optimization (GRPO) \cite{guo2025deepseek} as our base algorithm. GRPO optimizes policy $\pi_{\theta}$ by maximizing the following objective:
\begin{equation}
\begin{split}
    &\mathcal{J}(\theta) = \mathbb{E}[x \sim P(x), \mathcal{G} \sim \pi_{\theta_{\text{old}}}(y|x)] \\
    &\frac{1}{|\mathcal{G}|} \sum_{i=1}^{|\mathcal{G}|} \Big( 
    \min\Big(\frac{\pi_\theta(y_i|x)}{\pi_{\theta_{\text{old}}}(y_i|x)}A_i, \text{clip}\Big(\frac{\pi_\theta(y_i|x)}{\pi_{\theta_{\text{old}}}(y_i|x)}, 1-\epsilon, 1+\epsilon\Big)A_i\Big) - \beta\mathbb{D}_{\text{KL}}(\pi_\theta||\pi_{\text{ref}})\Big)
\end{split}
\label{eq::GRPO}
\end{equation}
where $A_i$ is the advantage computed using binary accuracy reward $r_i = \mathbb{I}(y_i = y)$ within $\mathcal{G}$:
\begin{equation}
    A_i = \frac{r_i - \text{mean}(\{r_1,r_2,\ldots,r_{|\mathcal{G}|}\})}{\text{std}(\{r_1,r_2,\ldots,r_{|\mathcal{G}|}\})}
\end{equation}

To evaluate different strategies for controlling reasoning length and efficiency, we implement the following baselines with different reward setting:

\textit{\textbf{(1) Length Controlled Policy Optimization (LCPO) }\cite{aggarwal2025l1controllinglongreasoning}}: it extends GRPO by modifying its reward function to include a correctness reward and a length penalty:
\begin{equation}
    r_i = \mathbb{I}(y_i = y) - \alpha \cdot |l_i - l^*|
\end{equation}
where $l_i$ is the generated output length, and $\alpha$ is a scalar that regulates the trade-off between generating the correct answer and meeting the target length $l^*$, which we set to 2,048 tokens during training.

\textit{\textbf{(2) Token Length Budget (TLB)} \cite{shen2025dastdifficultyadaptiveslowthinkinglarge}}: it establishes a token length budget metric $l_{b}$ based on sampling accuracy and length statistics, and designs a difficulty-adaptive reward as follows:
    \begin{equation}
    r_i = \begin{cases}
    \max(-0.5\lambda + 0.5, 0.1) & \text{if} \ \ y_i = y \\
    \min(0.9\lambda - 0.1, -0.1) & \text{otherwise}
    \end{cases}
\end{equation}

where $\lambda = l_i/l_{b} - 1$ denotes the relative length reward.

\textit{\textbf{(3) Cosine Reward Function (CosFn)} \cite{yeo2025demystifyinglongchainofthoughtreasoning}}: it introduces a cosine length scaling reward:
\begin{equation}
    r_i =  r_{\text{min}} + \frac{1}{2}(r_{\text{max}} - r_{\text{min}})(1 + \cos(\frac{l_i \ \pi}{l_{\text{max}}}))
\end{equation}
where different pairs of $(r_{\text{min}}, r_{\text{max}})$ are adopted from \cite{yeo2025demystifyinglongchainofthoughtreasoning}: $(0.5,1.0)$ for correct answers and $(-1,-0.5)$ for incorrect ones, with $l_{\text{max}}$ set to 2,048 tokens.

\paragraph{Evaluation Protocol}
We evaluate our method from three perspectives:

\textit{\textbf{(1) Math Accuracy.}} We adopt $\text{PASS}@1 = \frac{1}{|\mathcal{G}|}\sum_{i=1}^{|\mathcal{G}|} p_i$ as our primary accuracy metric following \cite{guo2025deepseek}, where $p_i$ indicates the correctness of the $i$-th response.

\textit{\textbf{(2) Reasoning Efficiency.}} We measure efficiency through two metrics: (1) the average token length $\overline{L} = \frac{1}{|\mathcal{G}|}\sum_{i=1}^{|\mathcal{G}|} l_i$, and (2) the average frequency of reflection words $\overline{n}_* = \frac{1}{|\mathcal{G}|}\sum_{i=1}^{|\mathcal{G}|} n_*(y_i)$ in $|\mathcal{G}|$ responses, where $n_* := \{ n_p, n_b\}$ represent the counts of ``Pause-Validation'' and ``Branch-Extension'' words, respectively.

\textit{\textbf{(3) Group Diversity.}} We quantify the diversity of $\mathcal{G}$ using the proposed entropy-based measures $H_l$, $H_p$, and $H_b$, which evaluate the distribution uniformity of response length, ``Pause-Validation'' words, and ``Branch-Extension'' words respectively.

\paragraph{Training Details and Hyperparameters}

For AdapThink training, we build upon the PEFT framework \cite{peft} using LoRA adaptation. Training uses a batch size of 32 and runs for 5 epochs with bfloat16 precision. The maximum token limit is set to 2K tokens during training, it is extended to 8K tokens for evaluation. For sampling, we generate 16 rollouts per question using temperature 0.7 and top-p 0.95, then select 8 samples based on our diversity-aware strategy with equal weights $\alpha_l = \alpha_b = \alpha_d = 1$. The confidence thresholds are set to $\varphi_{\text{low}} = 0.15$ and $\varphi_{\text{high}} = 0.5$. Following \cite{xiong2025minimalist}'s observation that total incorrect samples degrade GRPO training performance, we set a large $T_{\text{min}} = 3$ and $F_{\text{min}} = 1$ to preserve more correct samples for hard questions.

\section{Results}

In this section, we evaluate and analyze the performance of AdapThink across various settings through three key aspects: answer accuracy, reasoning efficiency, and group diversity.

\subsection{Overall Comparison} 
\begin{figure}[t]
\centering
\includegraphics[width=0.85\columnwidth]{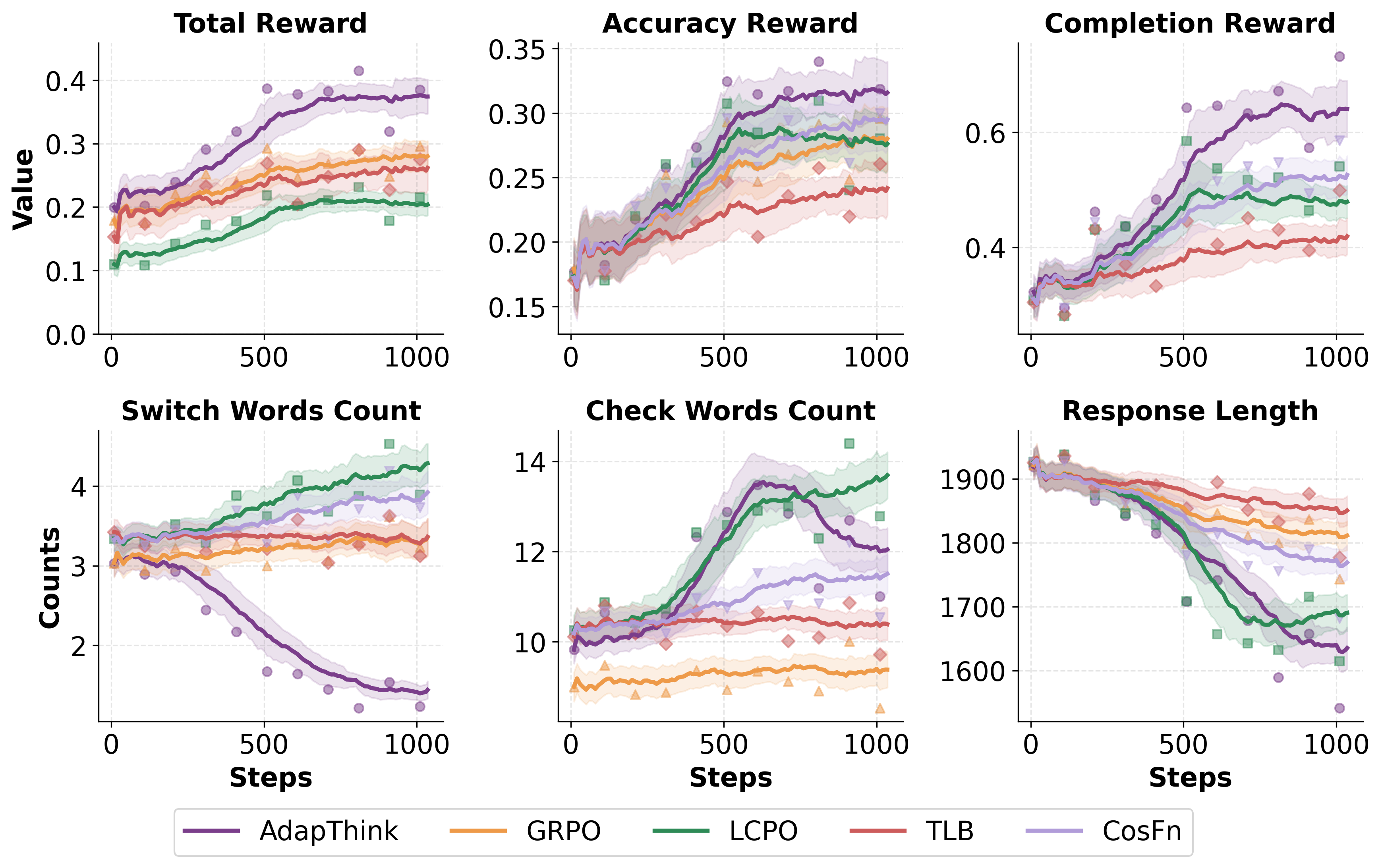}
\caption{Training comparison among five post-training frameworks on DeepSeek-Distill-Qwen-1.5B with a 2K token length constraint.}
\label{fig::train}
\end{figure}

As shown in Figure \ref{fig::train} and Table \ref{tab:main_results}, \emph{AdapThink significantly improves accuracy while reducing response length, demonstrating distinct patterns for different types of reflection words}. During training, AdapThink exhibits a notable decrease in switch words, while check words follow a pattern of initial increase followed by decline. This control over key reasoning words effectively generalizes to testing scenarios, resulting in a 27\% average performance improvement and 15.6\% reduction in response length compared to the base model.

For LCPO, we observe that the late-stage increase in check words does not correspond to improvements in completion or accuracy rewards. Interestingly, LCPO's strict length control shows a preference for shorter responses but fails to generalize to effective reasoning patterns, leading to an average 8.2\% lower accuracy compared to AdapThink during an 8K token limit.

For the GRPO and TLB methods, they demonstrated consistent minimal changes in reflection words during training. During the testing phase, their responses consumed an average of 215 and 488 more tokens per question respectively compared to AdapThink. While CosFn's separate length control for correct and incorrect answers leads to improved accuracy, its reasoning efficiency falls short of AdapThink.

\begin{table}[t]
\centering
\caption{Evaluation performance comparison across mathematical reasoning benchmarks with an 8K token length constraint.}
\label{tab:main_results}
\renewcommand{\arraystretch}{1.1}
\resizebox{\textwidth}{!}{
\begin{tabular}{l|l|cc|cc|cc|cc}
\toprule
\multirow{2}{*}{\textbf{Model}} & \multirow{2}{*}{\textbf{Method}} & 
\multicolumn{2}{c|}{\cellcolor{gray!10}\textbf{AIME2025}} & 
\multicolumn{2}{c|}{\cellcolor{gray!10}\textbf{AIME2024}} & 
\multicolumn{2}{c|}{\cellcolor{gray!10}\textbf{MATH500}} & 
\multicolumn{2}{c}{\cellcolor{gray!10}\textbf{AMC}} \\
\cmidrule(lr){3-4} \cmidrule(lr){5-6} \cmidrule(lr){7-8} \cmidrule(lr){9-10}
& & \textbf{PASS@1} & \textbf{Token} & \textbf{PASS@1} & \textbf{Token} & \textbf{PASS@1} & \textbf{Token} & \textbf{PASS@1} & \textbf{Token} \\
\midrule
\multirow{6}{*}{\makecell[c]{DeepSeek \\Distill Qwen \\ -1.5B-2K}} 
& Baseline & 17.92 & 7242 & 20.00 & 7451 & 77.25 & 3736 & 52.26 & 5743 \\
& \ \ +GRPO & 24.17\textcolor{deepred}{$\scriptstyle\uparrow{\scriptstyle 6.25}$} & 6763\textcolor{deepgreen}{$\scriptstyle\downarrow{\scriptstyle 479}$} & 26.67\textcolor{deepred}{$\scriptstyle\uparrow{\scriptstyle 6.67}$} & 6994\textcolor{deepgreen}{$\scriptstyle\downarrow{\scriptstyle 457}$} & 81.40\textcolor{deepred}{$\scriptstyle\uparrow{\scriptstyle 4.15}$} & 3090\textcolor{deepgreen}{$\scriptstyle\downarrow{\scriptstyle 646}$} & \textbf{62.95}\textcolor{deepred}{$\scriptstyle\uparrow{\scriptstyle 10.69}$} & 4839\textcolor{deepgreen}{$\scriptstyle\downarrow{\scriptstyle 904}$} \\
& \ \ +TLB & 20.83\textcolor{deepred}{$\scriptstyle\uparrow{\scriptstyle 2.91}$} & 6984\textcolor{deepgreen}{$\scriptstyle\downarrow{\scriptstyle 258}$} & 22.50\textcolor{deepred}{$\scriptstyle\uparrow{\scriptstyle 2.50}$} & 7200\textcolor{deepgreen}{$\scriptstyle\downarrow{\scriptstyle 251}$} & 80.40\textcolor{deepred}{$\scriptstyle\uparrow{\scriptstyle 3.15}$} & 3352\textcolor{deepgreen}{$\scriptstyle\downarrow{\scriptstyle 384}$} & 58.13\textcolor{deepred}{$\scriptstyle\uparrow{\scriptstyle 5.87}$} & 5239\textcolor{deepgreen}{$\scriptstyle\downarrow{\scriptstyle 504}$} \\
& \ \ +LCPO & 20.83\textcolor{deepred}{$\scriptstyle\uparrow{\scriptstyle 2.91}$} & 6724\textcolor{deepgreen}{$\scriptstyle\downarrow{\scriptstyle 518}$} & 29.17\textcolor{deepred}{$\scriptstyle\uparrow{\scriptstyle 9.17}$} & 6367\textcolor{deepgreen}{$\scriptstyle\downarrow{\scriptstyle 1084}$} & 78.60\textcolor{deepred}{$\scriptstyle\uparrow{\scriptstyle 1.35}$} & 2523\textcolor{deepgreen}{$\scriptstyle\downarrow{\scriptstyle 1213}$} & 61.14\textcolor{deepred}{$\scriptstyle\uparrow{\scriptstyle 8.88}$} & 4649\textcolor{deepgreen}{$\scriptstyle\downarrow{\scriptstyle 1094}$} \\
& \ \ +CosFn & 21.67\textcolor{deepred}{$\scriptstyle\uparrow{\scriptstyle 3.75}$} & 6956\textcolor{deepgreen}{$\scriptstyle\downarrow{\scriptstyle 286}$} & 28.33\textcolor{deepred}{$\scriptstyle\uparrow{\scriptstyle 8.33}$} & 7175\textcolor{deepgreen}{$\scriptstyle\downarrow{\scriptstyle 276}$} & 82.70\textcolor{deepred}{$\scriptstyle\uparrow{\scriptstyle 5.45}$} & 2729\textcolor{deepgreen}{$\scriptstyle\downarrow{\scriptstyle 1007}$} & 61.14\textcolor{deepred}{$\scriptstyle\uparrow{\scriptstyle 8.88}$} & 4847\textcolor{deepgreen}{$\scriptstyle\downarrow{\scriptstyle 896}$} \\
& \ \ +AdapThink & \textbf{25.42}\textcolor{deepred}{$\scriptstyle\uparrow{\scriptstyle 7.50}$} & 6455\textcolor{deepgreen}{$\scriptstyle\downarrow{\scriptstyle 787}$} & \textbf{31.67}\textcolor{deepred}{$\scriptstyle\uparrow{\scriptstyle 11.67}$} & 6812\textcolor{deepgreen}{$\scriptstyle\downarrow{\scriptstyle 639}$} & \textbf{82.80}\textcolor{deepred}{$\scriptstyle\uparrow{\scriptstyle 5.55}$} & 2742\textcolor{deepgreen}{$\scriptstyle\downarrow{\scriptstyle 994}$} & 62.24\textcolor{deepred}{$\scriptstyle\uparrow{\scriptstyle 9.98}$} & \textbf{4814}\textcolor{deepgreen}{$\scriptstyle\downarrow{\scriptstyle 929}$} \\
\midrule
\multicolumn{2}{l|}{FastCuRL-1.5B-24K\textsuperscript{‡}} & 25.42 & 7085 & 31.67 & 7117 & 86.40 & 3430 & 62.80 & 5440 \\
\multicolumn{2}{l|}{DeepScaler-1.5B-24K\textsuperscript{‡}} & 25.42 & 6397 & 37.50 & 6423 & 88.00 & 2831 & 66.87 & 4893 \\
\bottomrule
\multicolumn{10}{l}{\footnotesize \textsuperscript{‡}Originally trained with 24K token limit. We evaluate their official models with an 8K token limit.} 
\end{tabular}
}
\end{table}

\subsection{Ablation Studies}
To verify the effectiveness of each component in the AdapThink framework, we conducted three ablation experiments, and the evaluation results are shown in Table 2. 

\paragraph{Reward Components.} First, we performed ablation studies on the reward components $\lambda_l$ and $\lambda_b$ in GRPR. Without length control reward $\lambda_l$, we observed a substantial increase in pause-validation words, rising by 244.8\% for correct samples and 173.2\% for incorrect ones. The outputs became significantly longer, with $L^{\mathcal{T}}$ increasing by 17.7\% and a 2.1 percentage point accuracy drop. Similarly, removing $\lambda_b$ resulted in less efficient reasoning patterns with pause-validation frequency increasing by 65.5\% and branch-extension usage rising by 55.0\%, and a 2.5 percentage point drop in accuracy.

\paragraph{Diversity-aware Selection.} Subsequently, we investigated the effectiveness of the diversity-aware selection mechanism by setting different oversampling ratios $K$. Our experiments revealed that $K=2$ yielded better performance across all metrics, achieving 2.9\% higher accuracy with less token consumption. The higher diversity scores $H_l$ and $H_d$ suggest that the selection strategy promotes varied reasoning patterns in the training phase and accelerates the efficient exploration of RL.  

\paragraph{Curriculum Learning.} Inspired by \cite{deepscaler2025}, we implemented a curriculum learning for AdapThink. Specifically, we conducted secondary training on the best checkpoint from the 2K token limit to 4K. For comparison, we also trained a model directly with 4K token limit using AdapThink. 
Our results showed that progressive training from 2K to 4K token limit outperformed direct 4K training, achieving the highest accuracy while maintaining the most efficient reasoning patterns. 

\begin{table}[t]
\caption{Ablation studies of AdapThink. $L$ represents the maximum response token limit (in thousand) in training phase. The diversity metrics are from training phase while other metrics are from testing on AIME 2025. Superscripts $\mathcal{T}$ and $\mathcal{F}$ indicate correct and incorrect answers respectively.}
\label{tab:ablation}
\renewcommand{\arraystretch}{1.1}
\resizebox{\textwidth}{!}{
\begin{tabular}{ccccc|ccc|cccc|ccc}
\toprule
\multicolumn{5}{c|}{\textbf{Setting}} & 
\multicolumn{3}{c|}{\cellcolor{gray!10}\textbf{Accuracy}} & 
\multicolumn{4}{c|}{\cellcolor{gray!10}\textbf{Efficiency}} &
\multicolumn{3}{c}{\cellcolor{gray!10}\textbf{Diversity}} \\
\cmidrule(lr){1-5} \cmidrule(lr){6-8} \cmidrule(lr){9-12} \cmidrule(lr){13-15}
$\lambda_l$ & $\lambda_o$ & $\lambda_b$ & $K$ & $L$ & PASS@1 & $\overline{L}^{\mathcal{T}}$ & $\overline{L}^\mathcal{F}$ & $n_p^\mathcal{T}$ & $n_p^\mathcal{F}$ & $n_b^\mathcal{T}$ & $n_b^\mathcal{F}$ & $H_l$ & $H_b$ & $H_d$ \\
\midrule
\checkmark & \checkmark & \checkmark & 2 & 2 & 25.4 \textcolor{deepred}{$\scriptstyle\uparrow{\scriptstyle 7.5}$} & 3561 \textcolor{deepgreen}{$\scriptstyle\downarrow{\scriptstyle 580}$} & 7442 \textcolor{deepgreen}{$\scriptstyle\downarrow{\scriptstyle 478}$} & 16.5 \textcolor{deepgreen}{$\scriptstyle\downarrow{\scriptstyle 4.6}$} & \textbf{55.7} \textcolor{deepgreen}{$\scriptstyle\downarrow{\scriptstyle 9.4}$} & 2.0 \textcolor{deepgreen}{$\scriptstyle\downarrow{\scriptstyle 1.4}$} & 13.3 \textcolor{deepred}{$\scriptstyle\uparrow{\scriptstyle 1.1}$} & 25.7 \textcolor{deepred}{$\scriptstyle\uparrow{\scriptstyle 3.2}$} & 28.6 \textcolor{deepgreen}{$\scriptstyle\downarrow{\scriptstyle 5.3}$} & 35.5 \textcolor{deepred}{$\scriptstyle\uparrow{\scriptstyle 1.7}$} \\
\midrule
\ding{55} & \checkmark & \checkmark & 2 & 2 & 23.3 \textcolor{deepred}{$\scriptstyle\uparrow{\scriptstyle 5.4}$} & 4772 \textcolor{deepred}{$\scriptstyle\uparrow{\scriptstyle 630}$} & 8154 \textcolor{deepred}{$\scriptstyle\uparrow{\scriptstyle 235}$} & 56.9 \textcolor{deepred}{$\scriptstyle\uparrow{\scriptstyle 35.8}$} & 152.2 \textcolor{deepred}{$\scriptstyle\uparrow{\scriptstyle 87.1}$} & 3.4 \textcolor{deepgreen}{$\scriptstyle\downarrow{\scriptstyle 0.0}$} & \textbf{8.3} \textcolor{deepgreen}{$\scriptstyle\downarrow{\scriptstyle 3.9}$} & 22.2 \textcolor{deepgreen}{$\scriptstyle\downarrow{\scriptstyle 0.3}$} & 25.3 \textcolor{deepgreen}{$\scriptstyle\downarrow{\scriptstyle 8.6}$} & \textbf{38.9} \textcolor{deepred}{$\scriptstyle\uparrow{\scriptstyle 5.1}$} \\
\checkmark & \ding{55} & \checkmark & 2 & 2 & 22.5 \textcolor{deepred}{$\scriptstyle\uparrow{\scriptstyle 4.6}$} & 3775 \textcolor{deepgreen}{$\scriptstyle\downarrow{\scriptstyle 366}$} & 7853 \textcolor{deepred}{$\scriptstyle\uparrow{\scriptstyle 66}$} & 18.9 \textcolor{deepgreen}{$\scriptstyle\downarrow{\scriptstyle 2.2}$} & 71.4 \textcolor{deepred}{$\scriptstyle\uparrow{\scriptstyle 6.3}$} & 2.2 \textcolor{deepgreen}{$\scriptstyle\downarrow{\scriptstyle 1.2}$} & 12.2 \textcolor{deepgreen}{$\scriptstyle\downarrow{\scriptstyle 0.0}$} & 24.2 \textcolor{deepred}{$\scriptstyle\uparrow{\scriptstyle 1.7}$} & 27.8 \textcolor{deepgreen}{$\scriptstyle\downarrow{\scriptstyle 6.2}$} & 33.5 \textcolor{deepgreen}{$\scriptstyle\downarrow{\scriptstyle 0.4}$} \\
\checkmark & \checkmark & \ding{55} & 2 & 2 & 22.9 \textcolor{deepred}{$\scriptstyle\uparrow{\scriptstyle 5.0}$} & 4395 \textcolor{deepgreen}{$\scriptstyle\downarrow{\scriptstyle 253}$} & 7872 \textcolor{deepred}{$\scriptstyle\uparrow{\scriptstyle 48}$} & 27.3 \textcolor{deepred}{$\scriptstyle\uparrow{\scriptstyle 6.2}$} & 70.7 \textcolor{deepred}{$\scriptstyle\uparrow{\scriptstyle 5.6}$} & 3.1 \textcolor{deepred}{$\scriptstyle\uparrow{\scriptstyle 0.3}$} & 11.1 \textcolor{deepgreen}{$\scriptstyle\downarrow{\scriptstyle 1.1}$} & 23.5 \textcolor{deepred}{$\scriptstyle\uparrow{\scriptstyle 1.0}$} & 24.8 \textcolor{deepgreen}{$\scriptstyle\downarrow{\scriptstyle 9.2}$} & 33.8 \textcolor{deepgreen}{$\scriptstyle\downarrow{\scriptstyle 0.1}$} \\
\midrule
\checkmark & \checkmark & \checkmark & 1 & 2 & 22.5 \textcolor{deepred}{$\scriptstyle\uparrow{\scriptstyle 4.6}$} & 3947 \textcolor{deepgreen}{$\scriptstyle\downarrow{\scriptstyle 195}$} & 7793 \textcolor{deepred}{$\scriptstyle\uparrow{\scriptstyle 126}$} & 20.3 \textcolor{deepgreen}{$\scriptstyle\downarrow{\scriptstyle 0.8}$} & 71.1 \textcolor{deepred}{$\scriptstyle\uparrow{\scriptstyle 6.0}$} & 1.9 \textcolor{deepgreen}{$\scriptstyle\downarrow{\scriptstyle 1.5}$} & 10.9 \textcolor{deepgreen}{$\scriptstyle\downarrow{\scriptstyle 1.3}$} & 21.9 \textcolor{deepgreen}{$\scriptstyle\downarrow{\scriptstyle 0.6}$} & 26.5 \textcolor{deepgreen}{$\scriptstyle\downarrow{\scriptstyle 7.5}$} & 32.6 \textcolor{deepgreen}{$\scriptstyle\downarrow{\scriptstyle 1.2}$} \\
\checkmark & \checkmark & \checkmark & 1.75 & 2 & 25.0 \textcolor{deepred}{$\scriptstyle\uparrow{\scriptstyle 7.1}$} & 3803 \textcolor{deepgreen}{$\scriptstyle\downarrow{\scriptstyle 339}$} & 7579 \textcolor{deepgreen}{$\scriptstyle\downarrow{\scriptstyle 340}$} & 20.1 \textcolor{deepgreen}{$\scriptstyle\downarrow{\scriptstyle 1.0}$} & 58.8 \textcolor{deepgreen}{$\scriptstyle\downarrow{\scriptstyle 6.3}$} & 2.3 \textcolor{deepgreen}{$\scriptstyle\downarrow{\scriptstyle 1.1}$} & 12.6 \textcolor{deepred}{$\scriptstyle\uparrow{\scriptstyle 0.4}$} & 23.2 \textcolor{deepred}{$\scriptstyle\uparrow{\scriptstyle 0.8}$} & 32.6 \textcolor{deepgreen}{$\scriptstyle\downarrow{\scriptstyle 1.4}$} & 34.3 \textcolor{deepred}{$\scriptstyle\uparrow{\scriptstyle 0.5}$} \\
\midrule
\checkmark & \checkmark & \checkmark & 2 & 4 & 24.6 \textcolor{deepred}{$\scriptstyle\uparrow{\scriptstyle 6.7}$} & 3861 \textcolor{deepgreen}{$\scriptstyle\downarrow{\scriptstyle 281}$} & 7909 \textcolor{deepred}{$\scriptstyle\uparrow{\scriptstyle 10}$} & 23.9 \textcolor{deepred}{$\scriptstyle\uparrow{\scriptstyle 2.8}$} & 78.1 \textcolor{deepred}{$\scriptstyle\uparrow{\scriptstyle 13.0}$} & 3.4 \textcolor{deepgreen}{$\scriptstyle\downarrow{\scriptstyle 0.0}$} & 14.1 \textcolor{deepred}{$\scriptstyle\uparrow{\scriptstyle 1.9}$} & 24.5 \textcolor{deepred}{$\scriptstyle\uparrow{\scriptstyle 2.1}$} & 27.9 \textcolor{deepgreen}{$\scriptstyle\downarrow{\scriptstyle 6.1}$} & 26.8 \textcolor{deepgreen}{$\scriptstyle\downarrow{\scriptstyle 7.0}$} \\
\checkmark & \checkmark & \checkmark & 2 & 2$\rightarrow$4 & \textbf{26.0} \textcolor{deepred}{$\scriptstyle\uparrow{\scriptstyle 8.1}$} & \textbf{3339} \textcolor{deepgreen}{$\scriptstyle\downarrow{\scriptstyle 802}$} & \textbf{6886} \textcolor{deepgreen}{$\scriptstyle\downarrow{\scriptstyle 1034}$} & \textbf{13.8} \textcolor{deepgreen}{$\scriptstyle\downarrow{\scriptstyle 7.3}$} & 63.2 \textcolor{deepgreen}{$\scriptstyle\downarrow{\scriptstyle 1.9}$} & \textbf{1.7} \textcolor{deepgreen}{$\scriptstyle\downarrow{\scriptstyle 1.7}$} & 13.9 \textcolor{deepred}{$\scriptstyle\uparrow{\scriptstyle 1.7}$} & \textbf{31.6} \textcolor{deepred}{$\scriptstyle\uparrow{\scriptstyle 9.1}$} & \textbf{32.7} \textcolor{deepgreen}{$\scriptstyle\downarrow{\scriptstyle 1.3}$} & 27.9 \textcolor{deepgreen}{$\scriptstyle\downarrow{\scriptstyle 5.9}$} \\
\bottomrule
\end{tabular}
}
\end{table}

\section{Discussion}


\begin{figure}[t]
\centering
\includegraphics[width=\columnwidth]{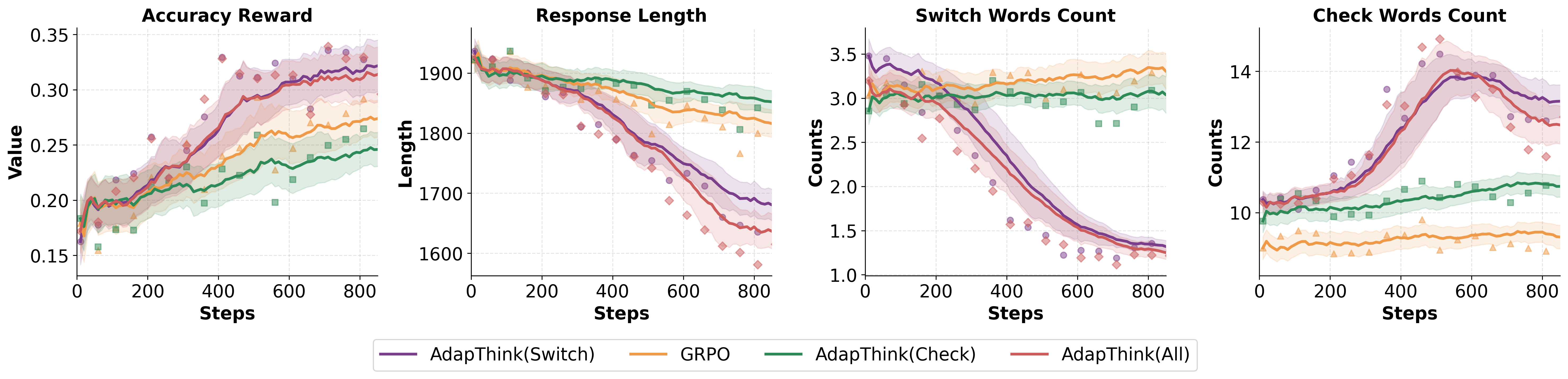}
\caption{Training comparison of AdapThink variants.}
\label{fig::discussion}
\end{figure}

\subsection{How Do Different Transition Words and Preferences Influence the Model's Performance?}

As shown in Figure \ref{fig::discussion}, we conducted a comparative analysis of three AdapThink variants by implementing control over different reflection words: AdapThink(check) calculates $\lambda_p$ in Equation \ref{eq::reason_reward} using the number of ``Pause-Validation'' words $n_p$, AdapThink(switch) controls ``Branch-Extension'' words $n_b$, and AdapThink(all) regulates both types simultaneously. Our experimental results demonstrate that reducing $n_p$ significantly influences learning efficiency, highlighting the crucial role of self-reflection processes even within short CoT reasoning. Additionally, AdapThink(all) exhibits a notable reduction in check words than AdapThink(switch), resulting in slightly degraded performance.

\subsection{Does Length-control Mechanisms Exhibit Reward Hacking Behavior?}

\begin{wraptable}{r}{0.4\textwidth}
\vskip -0.2in
\centering
\caption{N-gram repetition analysis across different post-training methods.}
\label{tab:ngram}
\renewcommand{\arraystretch}{1.1}
\begin{tabular}{l|ccc}
\toprule
\multirow{2}{*}{\textbf{Algorithm}} & \multicolumn{3}{c}{\textbf{N-grams (\%)}} \\
\cmidrule(lr){2-4}
& \textbf{Total} & $\mathcal{G_T}$ & $\mathcal{G_F}$ \\
\midrule
GRPO & \textbf{0.3} & \textbf{0.3} & \textbf{0.2} \\
TLB & 3.0 & 3.5 & 2.2 \\
LCPO & 10.8 & 5.9 & 19.8 \\
CosFn & 1.4 & 2.0 & \textbf{0.2} \\
AdapThink & 0.7 & 0.6 & 0.9 \\
\bottomrule
\end{tabular}
\vskip -0.1in
\end{wraptable}

To further investigate potential reward hacking introduced by length-control mechanisms, we employed N-gram (N=40) repetition rate metrics \cite{yeo2025demystifyinglongchainofthoughtreasoning} to quantify repetitive patterns in model responses for the MATH-500 test dataset, as shown in Table \ref{tab:ngram}. Our analysis reveals that LCPO exhibits severe repetitive patterns with an average N-gram repetition rate of 10.8\%, especially 19.8\% in incorrect answers, indicating potential reward hacking in its length control mechanism. While TLB and CosFn show moderate repetition, GRPO achieves the lowest overall repetition rate of 0.3\% as it avoids introducing dense rewards for length control. Notably, AdapThink maintains consistently low repetition rates across both correct and incorrect responses, suggesting its effectiveness in promoting rational reasoning patterns.

\section{Conclusion}
\label{sec::conclusion}
In this paper, we presented AdapThink, a novel RL post-training framework that adaptively modulates reasoning preferences based on model capabilities. We highlight two key insights from our study: (1) the effectiveness of dynamic reasoning control through group-relative rewards based on reflection words, completion markers and length, and (2) the importance of maintaining diversity in reasoning patterns during GRPR-based training.
Notably, with only a 2K token limit during training, our method outperformed existing baselines in both mathematical accuracy and token efficiency across various mathematical reasoning benchmarks.
One limitation and potential future direction lies in exploring semantic-level thought transitions control beyond current word-level control. We will further conduct experiments investigating the impact of non-reflection-related vocabulary on the reasoning process.

\newpage
\bibliographystyle{plain}
\bibliography{main}

\newpage
\appendix

\section{Examples of Model's Thinking Process}
\label{app::example}
\subsection{Example 1: Overthinking on a Simple Problem}
In this example, the model was presented with a straightforward probability problem. While it reached the correct answer early in its reasoning process (using only 479 tokens), it continued to reflect unnecessarily. The model's self-validation tendencies, marked by frequent use of phrases like ``Wait'', ``check'', and ``verify'', led to multiple alternative approaches and verifications. This resulted in consuming 2,613 tokens—more than five times the tokens needed for the initial correct solution.

\begin{examplebox}{Overthinking Example}
  \begin{questionbox}{Question}
    A bag contains 4 red balls and 3 blue balls. Two balls are drawn without replacement. What is the probability that both balls are of different colors?
  \end{questionbox}

  \begin{responsebox}{Model Response}
    \begin{thinkingbox}{Thinking Process}
    Okay, let's see. Well, I can think ...
    \textcolor{red}{\textbf{Alternatively}}, maybe I should ...
    \textcolor{red}{\textbf{Wait}}, is it right? ...
    \textcolor{red}{\textbf{Alternatively}}, maybe it is easier to ...
    \textcolor{mygreen}{So, the probability is 4/7}.
    
    But ... \textcolor{red}{\textbf{Alternatively}} ...
    But let me \textcolor{red}{\textbf{check}} both approaches ...
    \textcolor{red}{\textbf{Alternatively}} ... Approach 1 ...
    Wait ... Let me \textcolor{red}{\textbf{check}} ...
    Approach 2 ... \textcolor{red}{\textbf{Alternatively}} ...
    \textcolor{red}{\textbf{Wait}}, but let me just \textcolor{red}{\textbf{verify}} ...
    So all methods lead to 4/7. 
    
    \textcolor{mygreen}{**Final Answer**:} 
    
    The probability that both balls are of different colors is \boxed{\dfrac{4}{7}}.
    \end{thinkingbox}
    
    All methods confirm the probability that both balls are of different colors is \boxed{\dfrac{4}{7}}.
  \end{responsebox}

  \noindent\begin{minipage}{\textwidth}
    \hfill
    \begin{correctnessbox}{Correctness}
      Correct
    \end{correctnessbox}
    \hfill
    \begin{tokenbox}{Tokens Consumption}
      2613 (479 for first generating the correct answer.)
    \end{tokenbox}
    \hfill
    \begin{reflectionbox}{Reflection Words Count}
      19
    \end{reflectionbox}
    \hfill
  \end{minipage}
\end{examplebox}

\subsection{Example 2: Underthinking on a Complex Problem}
The second example shows the model attempting a more challenging modular arithmetic problem. Here, the model exhibited frequent but unproductive shifts in thinking, marked by numerous ``Wait'' and ``Alternatively'' statements. Unlike the first example where excessive reflection was merely inefficient, it reached the maximum token limit (4,096 token) before completing its calculation, resulting in an incorrect answer.

\begin{examplebox}{Underthinking Example}
  \begin{questionbox}{Question}
    Let $a_n=6^{n}+8^{n}$. Determine the remainder upon dividing $a_ {83}$ 
by $49$?
  \end{questionbox}

  \begin{responsebox}{Model Response}
    \begin{thinkingbox}{Thinking Process}
    Okay, so I need to find the remainder ...
    Let me think ... First ...
    But \textcolor{red}{\textbf{wait}} ...
    \textcolor{red}{\textbf{Alternatively}} ... \textcolor{red}{\textbf{Alternatively}} ...
    \textcolor{red}{\textbf{Wait}} ... Let's \textcolor{red}{\textbf{check}}... \textcolor{red}{\textbf{Wait}}, perhaps there's a better way ...
    Oh! \textcolor{red}{\textbf{Wait}} a second ...
    \textcolor{red}{\textbf{Wait}} ... Okay, so that's the first part ...
    \textcolor{red}{\textbf{Wait}} ... \textcolor{red}{\textbf{Wait}} first ...
    \textcolor{red}{\textbf{Wait}}, perhaps ...
    \textcolor{red}{\textbf{Wait}}, but just to make sure ...
    Similarly,  $8^7 \equiv 1 \pmod{13}$, 83 divided by 7 gives 11*7=
    \end{thinkingbox}
  \end{responsebox}

  \noindent\begin{minipage}{\textwidth}
    \hfill
    \begin{incorrectbox}{Correctness}
      Incorrect
    \end{incorrectbox}
    \hfill
    \begin{tokenbox}{Tokens Consumption}
      4096 (max token limit)
    \end{tokenbox}
    \hfill
    \begin{reflectionbox}{Reflection Words Count}
      35
    \end{reflectionbox}
    \hfill
  \end{minipage}
\end{examplebox}

\section{Distribution of Reflection Words in Model Responses}

\begin{figure}[t]
\centering
\includegraphics[width=\textwidth]{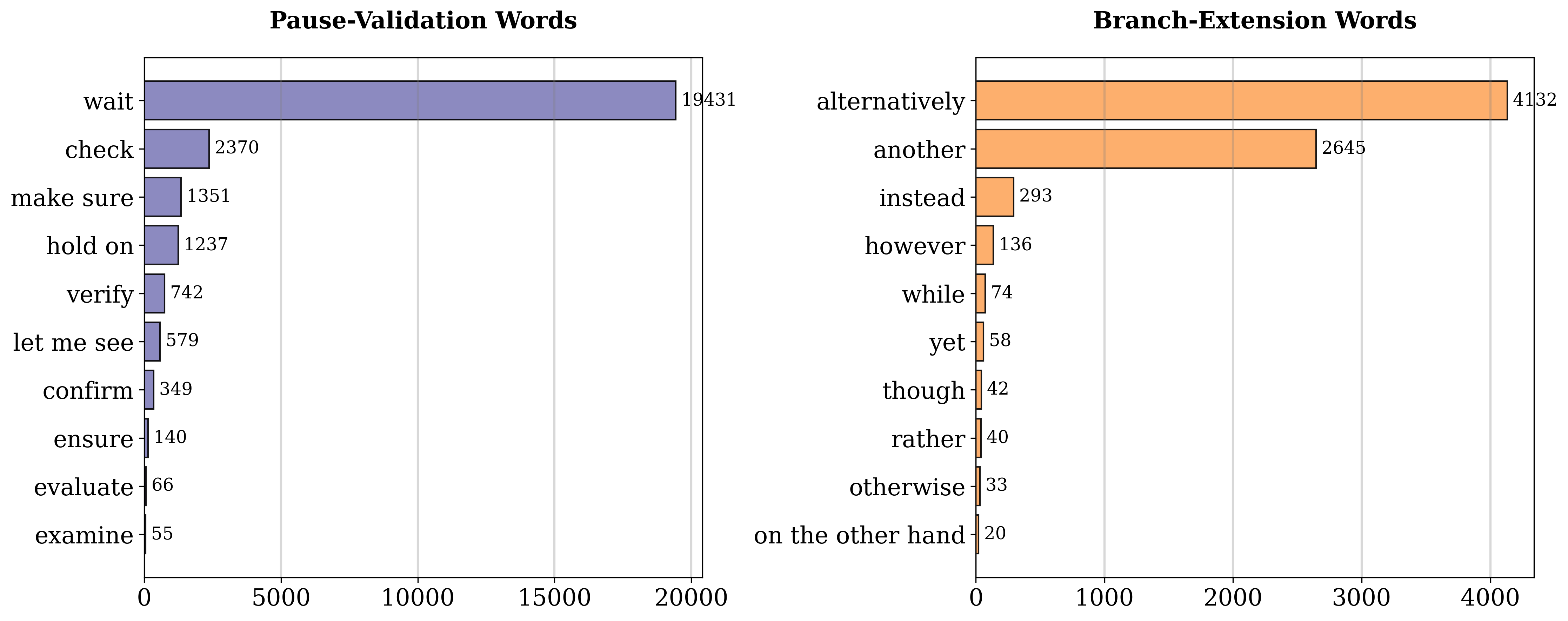}
\caption{Frequency distribution of reflection words in base model's responses.}
\label{fig:reflection_words}
\end{figure}

To provide a comprehensive understanding of our choice of reflection words in Section \ref{sec::observation}, we conducted a detailed frequency analysis of various transition words in Deepseek-distilled Qwen 1.5B model's responses. Figure \ref{fig:reflection_words} presents the complete distribution of both "Pause-Validation" and "Branch-Extension" words.

\section{AdapThink Algorithm Details}
\label{app::pseudo-code}
As introduced in Section \ref{sec::methodology}, AdapThink proposes a novel post-training framework that adaptively controls model's reasoning behavior through group-level response characteristics. Here we provide the complete algorithm details of AdapThink (Algorithm \ref{alg:adapthink}).

\begin{algorithm}
\caption{AdapThink: Adaptive Thinking Post-trainig Framework}
\label{alg:adapthink}
\begin{algorithmic}[1]
\Require Pre-trained model $\pi_{\theta}$, Dataset $\mathcal{D}$, Oversample factor $K$; Confidence thresholds $\varphi_{\text{low}}$, $\varphi_{\text{high}}$;Minimum correct/incorrect samples $T_{\min}$, $F_{\min}$
\For{each $(x, y) \in \mathcal{D}$}
    \State Generate $K|\mathcal{G}|$ samples $\mathcal{G}'$ using $\pi_{\theta}$
    \State Compute model confidence $\varphi$ using Equation \ref{eq::confidence}
    \State Partition $\mathcal{G}'$ into $\mathcal{G_T}'$ and $\mathcal{G_F}'$ based on correctness
    \State $T \leftarrow \min\{|\mathcal{G_T}'|, T_{\min}\}$, $F \leftarrow \min\{|\mathcal{G_F}'|, F_{\min}\}$
    \If{$\varphi \leq \varphi_{\text{low}}$}
        \State Select $T$ correct samples from $\mathcal{G_T}'$ maximizing $H_{\text{tot}}$
        \State Select remaining samples from $\mathcal{G}' \setminus \mathcal{G_T}'$ maximizing $H_{\text{tot}}$
    \Else
        \State Select $F$ incorrect samples from $\mathcal{G_F}'$ maximizing $H_{\text{tot}}$
        \State Select remaining samples from $\mathcal{G}' \setminus \mathcal{G_F}'$ maximizing $H_{\text{tot}}$
    \EndIf
    \State Compute component rewards $\lambda_l$, $\lambda_b$, $\lambda_o$ using Equation \ref{eq::component_reward}
    \State Calculate adaptive weight $\omega(\varphi)$ using Equation \ref{eq::weight}
    \State Update $\pi_{\theta}$ using Equation \ref{eq::GRPO} with
    GRPR reward (Equation \ref{eq::reason_reward}) and accuracy reward
\EndFor
\\
\Return Updated model $\pi_{\theta}$
\end{algorithmic}
\end{algorithm}

\section{Hyperparameter Configuration}
The core hyperparameters used in AdapThink are summarized in Table~\ref{table:hyperparameters}. The training is on 8 NVIDIA H100 GPUs, taking approximately 22 hours to complete 1000 steps.

\begin{figure}[ht]
\centering
\includegraphics[width=\textwidth]{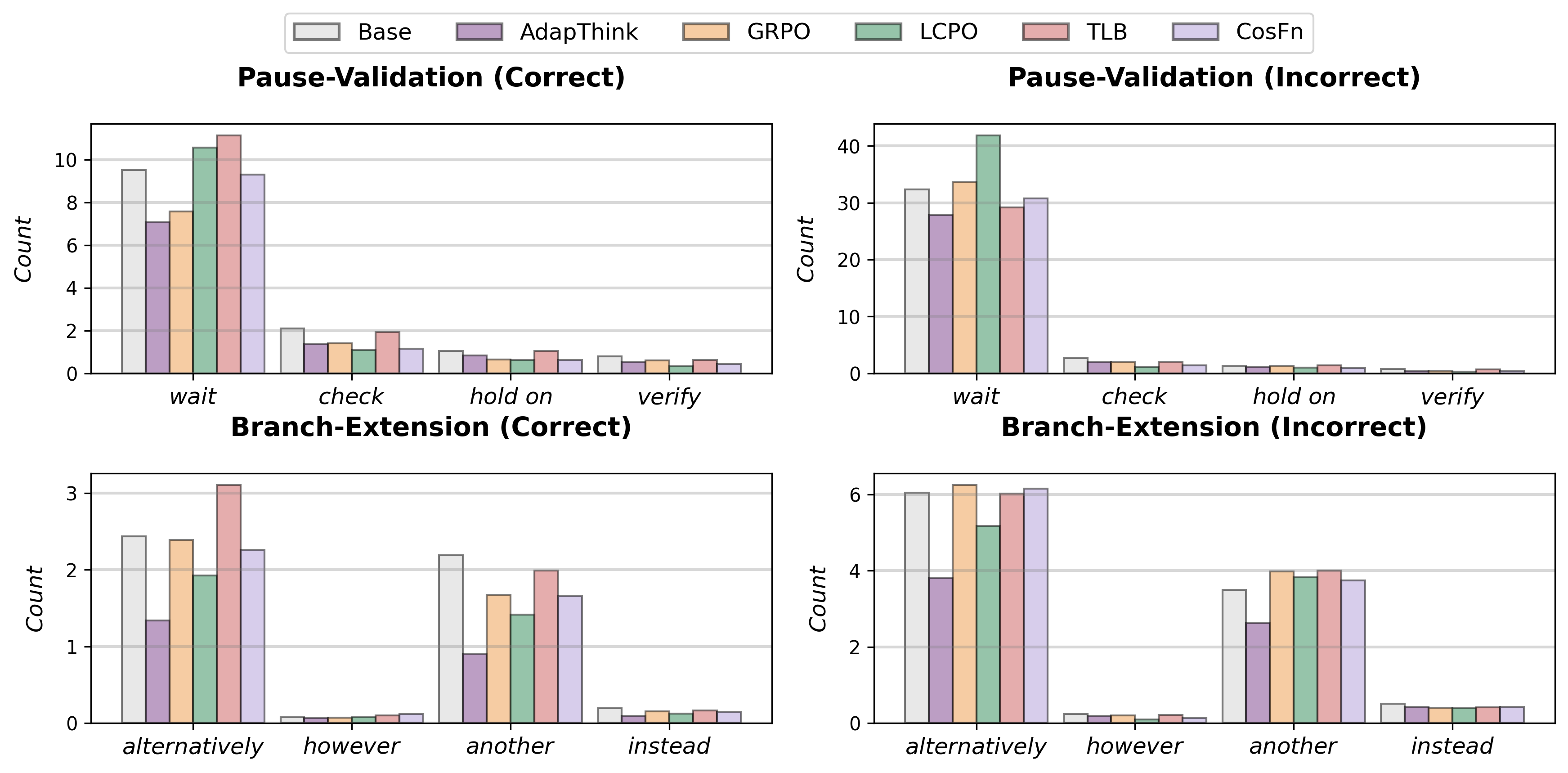}
\caption{Distribution of reflection words across different post-training methods for correct and incorrect responses in MATH-500 datasets.}
\label{fig:test_reflection_words}
\end{figure}

\begin{table}[ht]
\centering
\resizebox{\textwidth}{!}{
\begin{tabular}{lll}
\toprule
\textbf{Category} & \textbf{Parameter} & \textbf{Value} \\
\midrule
\multirow{4}{*}{Model Settings} & Base Model & DeepSeek-R1-Distill-Qwen-1.5B \\
 & Torch Dtype & bfloat16 \\
 & Max Sequence Length & 1024 \\
 & Max New Tokens & 2048 \\
\midrule
\multirow{5}{*}{Training Settings} & Learning Rate & 2e-6 \\
 & Number of Epochs & 5 \\
 & Batch Size & 8 \\
 & Gradient Accumulation Steps & 8 \\
 & Warmup Steps & 20 \\
\midrule
\multirow{2}{*}{LoRA Settings} & LoRA Rank (r) & 32 \\
 & LoRA Alpha & 32 \\
\midrule
\multirow{6}{*}{RL Settings} & Algorithm & GRPO \\
 & KL Coefficient & 0.15 \\
 & Number of Generations & 16 \\
 & Target Generations & 8 \\
 & Min Correct Generations & 3 \\
 & Min Incorrect Generations & 1 \\
\midrule
\multirow{2}{*}{Generation Settings} & Temperature & 0.7 \\
 & Top-p & 0.95 \\
\bottomrule
\end{tabular}
}
\caption{Core hyperparameters for AdapThink post-training configuration}
\label{table:hyperparameters}
\end{table}

\section{More Results}

\subsection{Analysis of Detailed Reflection Words Distribution}
\label{appendix:test_reflection_words}

To provide deeper insights into \textit{how different post-training methods affect the model's reasoning behavior}, we analyze the distribution of reflection words across correct and incorrect responses, as shown in Figure~\ref{fig:test_reflection_words}. 

For most pause-validation words and branch-extension words, we observe that AdapThink significantly reduces the frequency of validation words in both correct and incorrect responses compared to the base model, suggesting a more efficient reasoning process. In contrast, GRPO and TLB exhibit less pronounced changes in these reflection words; LCPO shows elevated counts of pause words, particularly in incorrect responses.

Our analysis provides insights into the relationship between reflection words and model performance. While LCPO achieves the shortest response lengths in Table. \ref{tab:main_results}, its elevated use of pause words, especially in incorrect responses, suggests redundant reasoning patterns that may explain its relatively lower accuracy improvements. In contrast, AdapThink's systematic reduction in reflection words contributes to more efficient token usage while maintaining effective reasoning paths, resulting in consistent accuracy improvements across all benchmarks within token constraints. 

\subsection{Analysis of More Ablation}
\label{appendix:training_dynamics}

We conducted additional ablation studies to explore two variants:

\begin{itemize}
    \item \textit{AdapThink(Sequential)} - where we calculate $\lambda_b$ based on the number of sequential words (``first'',``then'', ``next'', ``finally'', ``therefore'', ``so'', ``thus'') within each group.
    \item \textit{AdapThink(No weight)} - where we remove the weighting mechanism for rewards, consistently favoring shorter responses with minimal reflection.
\end{itemize}

\begin{figure}[ht]
\centering
\includegraphics[width=\textwidth]{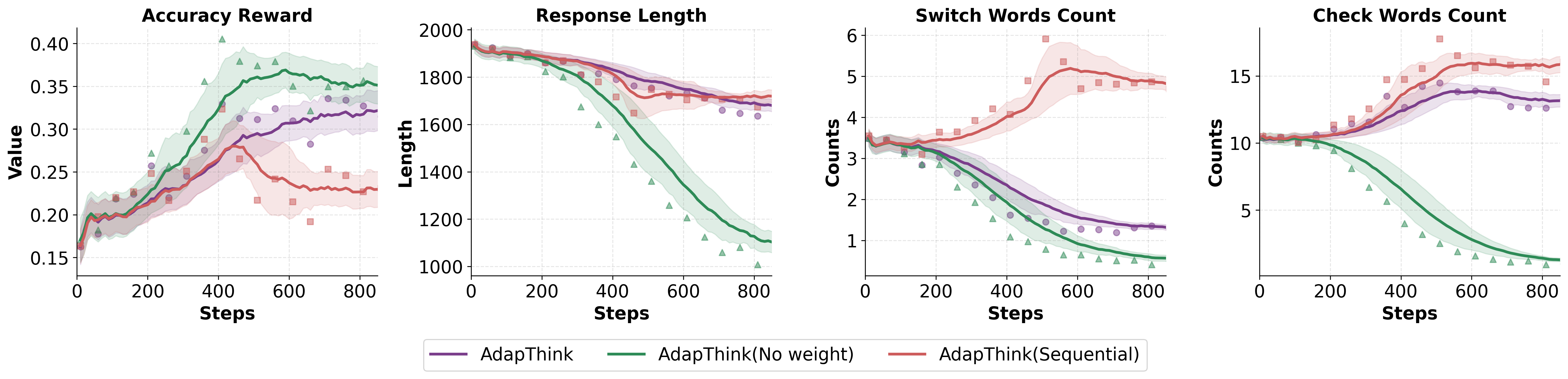}
\caption{Training dynamics comparison among three AdapThink variants.}
\label{fig:training_dynamics}
\end{figure}

We present comparative results between these variants and the original AdapThink model, both during training comparison in Figure \ref{fig:training_dynamics} and testing on the AIME2025 benchmark in Figure \ref{fig:training_dynamics}.

\begin{figure}[ht]
\centering
\includegraphics[width=\textwidth]{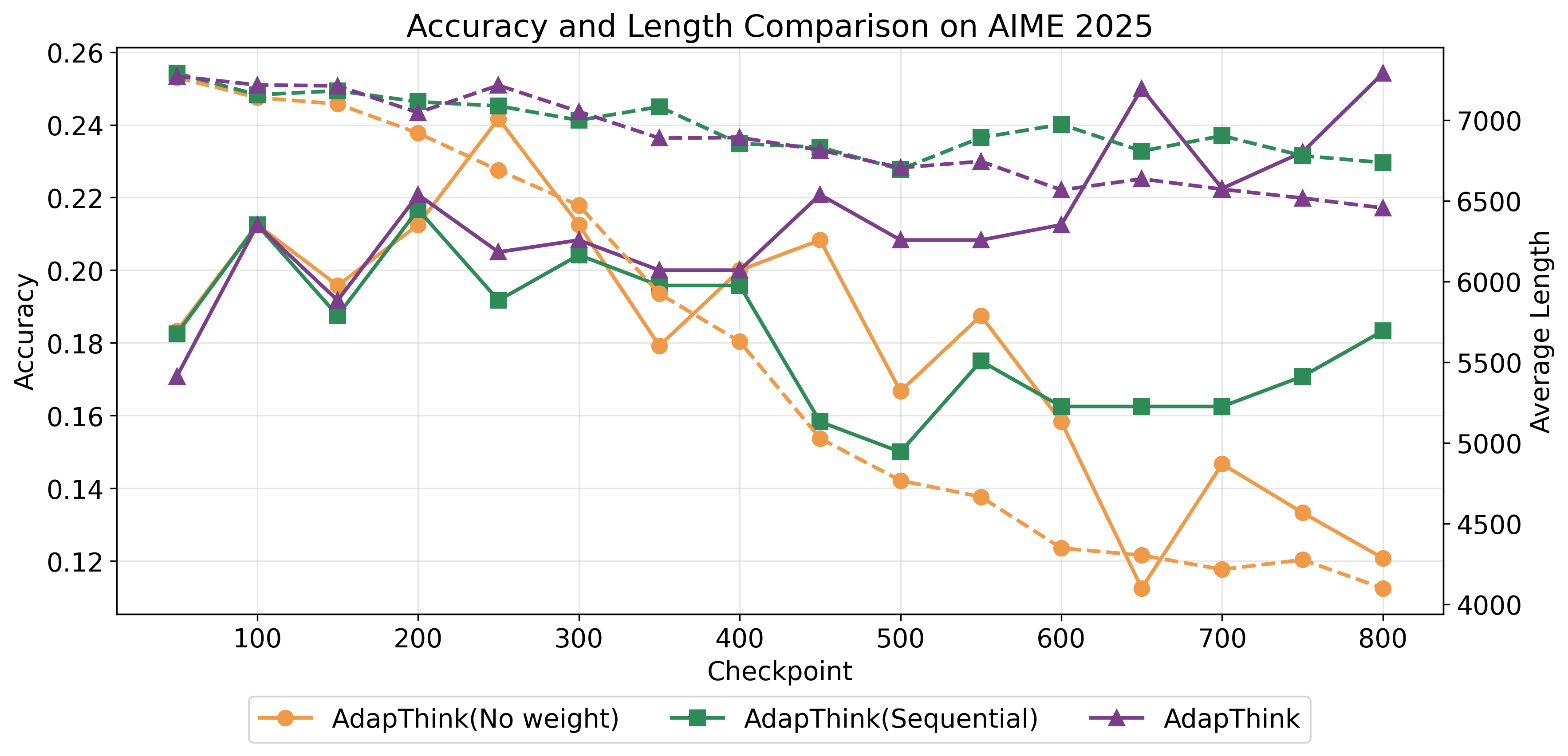}
\caption{Test comparison among three AdapThink variants across checkpoints at 50-step intervals.}
\label{fig:ablation_test}
\end{figure}

Interestingly, we observed two key findings:

(1) AdapThink(No weight) significantly reduced response lengths, which initially appeared advantageous in the early stages of post-training due to the inherent redundancy of reflection words in the base model. However, this eventually led to decreased accuracy rewards on the training set, with even more pronounced deterioration on the more challenging AIME 2025 dataset. As shown in Figure \ref{fig:ablation_test}, we found that the reduced length in later stages of AdapThink(No weight) corresponded to lower PASS@1, highlighting a limitation of existing length control algorithms. Adapthink's dynamic weighting mechanism based on model capability mitigates this issue.

(2) AdapThink(Sequential) showed no significant changes in either performance or length. Our observations during both training and testing revealed that it consistently produced longer outputs compared to the original AdapThink algorithm, suggesting that controlling sequential words does not effectively address the overthinking/underthinking phenomenon.

\subsection{Example of AdapThink and Base Model}
\begin{examplebox}{AdapThink Example (from AIME 2025)}
  \begin{questionbox}{Question}
    Find the number of ordered pairs $(x,y)$, where both $x$ and $y$ are integers between $-100$ and $100$, inclusive, such that $12x^{2}-xy-6y^{2}=0$.
  \end{questionbox}

  \begin{responsebox}{Model Response}
    \begin{thinkingbox}{Thinking Process}
    Okay, so I need to find the number of ordered pairs $(x, y)$ where both $x$ and $y$ are integers between -100 and 100, inclusive. The condition given is $12x^2 - xy - 6y^2 = 0$. Hmm, let me try to figure this out step by step. 
    
    First, looking at the equation: $12x^2 - xy - 6y^2 = 0$. It seems like a quadratic in terms of x and y. Maybe I can factor it or rearrange it somehow. ... Let's see:    
    $$12x^2 - xy - 6y^2 = 0$$
    Divide both sides by 3:   
    $$4x^2 - \frac{xy}{3} - 2y^2 = 0$$
    
    \textcolor{red}{\textbf{Wait}}, that doesn't help much because we have a fraction. Maybe factoring differently.  ...
    
    So, seems consistent.  Therefore, the number of ordered pairs is 117.
    \textcolor{mygreen}{**Final Answer**:} 
    The number of ordered pairs is \boxed{117}.    
    \end{thinkingbox}
    To find the number of ordered pairs $(x, y)$ where both $x$ and $y$ are integers between -100 and 100, inclusive, such that $12x^2 - xy - 6y^2 = 0$, we start by ... Thus, we subtract 1 to avoid double-counting.  
    The total number of ordered pairs is:\[ 51 + 67 - 1 = 117 \]  Thus, the number of ordered pairs is \boxed{117}.
  \end{responsebox}

  \noindent\begin{minipage}{\textwidth}
    \hfill
    \begin{correctnessbox}{Correctness}
      Correct
    \end{correctnessbox}
    \hfill
    \begin{tokenbox}{Tokens Consumption}
      \textbf{4502}
    \end{tokenbox}
    \hfill
    \begin{reflectionbox}{Reflection Words Count}
      \textbf{19} (\textbf{1} ``Branch-Extension'' word and \textbf{18} ``Pause-Validation'' Words)
    \end{reflectionbox}
    \hfill
  \end{minipage}
\end{examplebox}

\begin{examplebox}{DeepSeek-distilled Qwen 1.5B Example (from AIME 2025)}
  \begin{questionbox}{Question}
    Find the number of ordered pairs $(x,y)$, where both $x$ and $y$ are integers between $-100$ and $100$, inclusive, such that $12x^{2}-xy-6y^{2}=0$.
  \end{questionbox}

  \begin{responsebox}{Model Response}
    \begin{thinkingbox}{Thinking Process}
    Alright, so I have this problem here: I need to find the number of ordered pairs $(x, y)$ where both $x$ and $y$ are integers between $-100$ and $100$, inclusive. The equation given is $12x^2 - xy - 6y^2 = 0$. Hmm, okay. Let me try to figure this out step by step. 
    
    First, I notice that this is a quadratic equation in terms of x and y. Maybe I can factor it or rewrite it in a way that makes it easier to solve. Let me write it down again:

    $$12x^2 - xy - 6y^2 = 0$$

    \textcolor{red}{\textbf{Alternatively}}, maybe factor the quadratic expression as $(ax + by)(cx + dy) = 0$. Let me try that...

    \textcolor{red}{\textbf{Alternatively}}, perhaps I can write the equation as: $12x^2 -xy -6y^2=0$...

    \textcolor{red}{\textbf{Alternatively}}, maybe factor it as $(ax + by)(cx + dy)=0$...
    
    \textcolor{red}{\textbf{Alternatively}}, perhaps factor out a 3: $$3(4x^2 - \frac{1}{3}xy -2y^2)=0$$

    But that doesn't seem helpful.
    
    \textcolor{red}{\textbf{Alternatively}}, maybe factor as $(4x + ay)(3x + by)=0$ ...

    So, I think I can confidently say the answer is 117.
    \textcolor{mygreen}{**Final Answer**:} 
    \boxed{117}.    
    \end{thinkingbox}
    We need to find the number of ordered pairs $(x, y)$ where both $x$ and $y$ are integers between $-100$ and $100$, inclusive, such that $12x^2 - xy - 6y^2 = 0$ ...
    The total number of solutions is: $$67 + 51 - 1 = 117$$ Thus, the number of ordered pairs $(x, y)$ is: \boxed{117},
  \end{responsebox}

  \noindent\begin{minipage}{\textwidth}
    \hfill
    \begin{correctnessbox}{Correctness}
      Correct
    \end{correctnessbox}
    \hfill
    \begin{tokenbox}{Tokens Consumption}
      \textbf{7377}
    \end{tokenbox}
    \hfill
    \begin{reflectionbox}{Reflection Words Count}
      \textbf{46} (\textbf{10} ``Branch-Extension'' words and \textbf{36} ``Pause-Validation'' Words)
    \end{reflectionbox}
    \hfill
  \end{minipage}
\end{examplebox}

\end{document}